# Nonnegative Matrix Factorization applied to reordered pixels of single images based on patches to achieve structured nonnegative dictionaries☆

Richard M. Charles[a,∗], Kye M. Taylor[b,], James H. Curry[a]

[a]*University of Colorado at Boulder, Department of Applied Mathematics, Box 526 UCB, Boulder, CO, 80309-0526, United States*
[b]*Tufts University, Department of Mathematics, Broomfield-Pearson, 503 Boston Ave., Medford, MA, 02155, United States*

**Abstract**

Recent improvements in computing allow for the processing and analysis of very large datasets in a variety of fields. Often the analysis requires the creation of low-rank approximations to the datasets leading to efficient storage. This article presents and analyzes a novel approach for creating nonnegative, structured dictionaries using NMF applied to reordered pixels of single, natural images. We reorder the pixels based on patches and present our approach in general. We investigate our approach when using the Singular Value Decomposition (SVD) and Nonnegative Matrix Factorizations (NMF) as low-rank approximations. Peak Signal-to-Noise Ratio (PSNR) and Mean Structural Similarity Index (MSSIM) are used to evaluate the algorithm. We report that while the SVD provides the best reconstructions, its dictionary of vectors lose both the sign structure of the original image and details of localized image content. In contrast, the dictionaries produced using NMF preserves the sign structure of the original image matrix and offer a nonnegative, parts-based dictionary.

---

☆This document is a collaborative effort between researchers at CU Boulder and Tufts University
∗Principal corresponding author
*Email addresses:* `richard.charles@colorado.edu`, 1(720)273 4408(p), 1(303)492 4066(f) (Richard M. Charles), `kye.taylor@tufts.edu` (Kye M. Taylor), `James.H.Curry@colorado.edu` (James H. Curry)

1. **Introduction**

The notion of *patches* is increasingly popular in image processing research. Patches are local portions, or snippets, of an image. Applications that have benefited from patch-based algorithms include denoising [3, 10], inpainting [10], feature extraction [12, 14], and compression [1, 11, 9, 2]. Image compression algorithms seek to find redundancies within images in one of three ways: 1) coding redundancy, 2) spatial redundancy and 3) irrelevant or superfluous information. If given an input image, compression algorithms leverage these redundancies to encode and then decode the data. Thus, we propose a lossy compression algorithm leveraging the low-rank approximation schemes of Nonnegative Matrix Factorization (NMF) and Singular Value Decomposition (SVD) applied to patches of an image. When using NMF on a single image, our approach successfully achieves a dictionary of vectors which is nonnegative and reflects structure similar to the patches of the original image. The novelty of this approach lies in the fact that NMF is applied to a library of patches associated with a single image for the purpose of compression. We also provide a methodology for rearranging the entries of a matrix in order to achieve higher compression rates. The approach involves an encoding scheme based on reordered pixels of the original image and achieves better compression rates compared to its application on the columns of unaltered images.

If we regard an intensity image as an $n \times m$ matrix, $A$, then a patch is equivalent to a submatrix of $A$. Algorithms that are based on submatrices, or *blocks*, are designed primarily to parallelize or optimize computations of matrix operations [16, 6]. In contrast, patch-based algorithms are motivated by the fact that the set of patches provides a mechanism to emphasize and upncover structure in the local correlation of pixels and repeating patterns/textures that are characteristic of natural images.

In this article, we present and analyze a patch-based algorithm for generating low-rank approximations to an image and creating nonnegative, structured dictionaries. That is, given an algorithm that returns a low-rank approximation of an input matrix, we add a pre-processing step that amounts to reordering pixels in the input image based on a set of patches. These patches are akin to the images used originally in the application of NMF. We then apply the low-rank approximation algorithm to this reordered version of the image, which can be used as a compressed, or encoded, representation of the original input image. Given the encoded image, the decoded image, or reconstruction, is obtained by inverting the reordering process. We use the PSNR and SSIM as measures for comparisons of compressed images quality of various ranks and patch sizes.

We note that our approach is similar to the algorithm presented in [11]. However, we expand on the the work in [11] by considering a wider range of patch sizes, more general low-rank approximations and the nature of the dictionaries which are generated. Specifically, while [11] considers a low-rank approximation strategy based only on matrix singular value decomposition (SVD) [16, 6], we present a generalized framework, and provide experiments that are based on nonnegative matrix factorization (NMF) [8], in addition to the SVD. Although NMF has been used prior for the purpose of compressing libraries of images, this is the first time it is being used on patches of a single image for the purpose of generating low-rank approximations. The generalization we describe sug-gests that each patch should be organized as a column in the reordered matrix, whereas [11] organizes each patch as a row. Finally, we also provide a discussion on the computational advantage that can be achieved when using smaller patches than those considered in [11].

The computational advantage that can be achieved is based on the simple observation that if we take an $n \times m$ intensity image represented as a matrix, and organize it into another matrix that is $p^2 \times \frac{nm}{p^2}$, then for small enough values of $p$, the maximal rank of the reordered matrix is $p^2$. Therefore, the reordered matrix can be well-approximated with a matrix of lower rank than the original matrix, and therefore can be computed more efficiently. Hence this provokes the question, does reordering the original image cost too much? Fortunately, we show that it is computationally inexpensive compared to *not* reordering the original image matrix and instead proceeding with the traditional approach of computing a low-rank approximation to the original $n \times m$ image. We expand on this observation in Section 5.1.

### 1.1. The role of NMF in the proposed algorithm

We point out that NMF yields a low-rank approximation to a matrix, akin to the SVD. However, unlike the SVD, NMF has the additional features of preserving the sign structure of the input matrix, as well as producing a matrix factorization comprising a matrix of nonnegative weights and another matrix with columns that we can interpret as nonnegative components, or patches, that can be used to produce an approximation to the original input image [8]. Hence, we consider the NMF for the following three reasons: 1) it is capable of providing a low-rank approximation of an image, 2) it preserves the sign-structure of the original dataset, and 3) it produces a *parts-based decomposition* of the input image. We detail this interpretation and explore the parts that are recovered by NMF from test images in Sections 3.2.1 and 4.1.1.

## 1.2. Outline

This paper is organized as follows. In Section 2, we begin by describing background information related to computing the low-rank approximations that are used on the reordered version of the image matrix in the proposed algorithm. Further, we also compute the low-rank approximations of the original images for comparison with the proposed algorithm. We detailed the proposed algorithm in Section 3 where we will identify the distinction between the low-rank approximation of the reordered matrix that is used in our algorithm, and the low-rank approximation of the original matrix. Section 4 provides qualitative and quantitative assessments of the reconstruction quality and compression rates that can be achieved with the proposed algorithm. We show performance curves and investigate the optimum patch size based on the PSNR of the reconstructions obtained experimentally using both SVD and NMF in 4.3. We investigate the performance of the algorithm using MSSIM as an additional measure of the quality of the approximations and provide details in section 4.3. We provide the dictionary of vectors which are generated using NMF, in the form of patches in order to investigate the structures which are revealed. We then conclude with a discussion on the computational efficiency that can be achieved and end by presenting several open questions in Section 5.

## 2. Setup

### 2.1. Preliminaries

Let $A \in \mathbb{R}^{n \times m}$ represent an $n \times m$ pixel intensity image. A common approach to achieving a low-rank approximation to the image is to decompose the matrix $A$ as a product of matrices. For example, a goal is to find $W \in \mathbb{R}^{n \times k}$ and $H \in \mathbb{R}^{k \times m}$ for which

$$A \approx W H. \qquad (1)$$

Assuming that $W$ and $H$ are full rank, the right-hand-side of (1) represents a *rank-k approximation* to $A$, where each column of $A$ is assumed to be a linear combination of columns of $W$. Examples of matrix decompositions that fit the form in (1) are the singular value decomposition (SVD) (which is related to principal component analysis [5]), nonnegative matrix factorization (NMF), as well as vector quantization. The difference between each of these decomposition lies in the constraints imposed on $W$ and $H$ [8].

## 2.2. Memory footprint

With the approximation (1) in hand, compression can be achieved by either encoding the matrix $H$ or both $W$ and $H$. In this article, we will assume that both $W$ and $H$ are stored as part of the encoding. For the sake of simplicity, rather than present a bit-allocation strategy for storing this information, we measure the amount of information used in a compression strategy with the total number of values, or elements, used to define a matrix approximation. We refer to this number as the *memory footprint* of a compressed matrix. In particular, since $A$ has $nm$ entries, we say the memory footprint of $A$ is $nm$.[1] Similarly, the memory footprint of the *encoded* image on the right of (1) is $k(n+m)$. Therefore, the integer $k$ is chosen so that $k(n+m) < nm$ in order to achieve compression.

---

[1] We do not count the memory required to store the $nm$ additional values for the rank $k$ SVD.

The *decoded* image can be defined by the operator $T_k : \mathbb{R}^{n \times m} \to \mathbb{R}^{n \times m}$, which returns the rank-$k$ approximation to $A$ given in (1). That is

$$T_k(A) = W H. \tag{2}$$

*2.3. SVD and NMF for low-rank approximations*

Below, we describe how the SVD and NMF are used to compute $T_k(A)$ in (2). The use of these decompositions in the proposed algorithm are detailed in Section 3.2. We also compare the two approximations in more detail in Section 5.

*2.3.1. Low-rank approximations via the SVD*

To compute the factorization (2) with the SVD, we proceed in the standard fashion, and write the SVD of the matrix $A \in \mathbb{R}^{n \times m}$ as

$$A = USV^T,$$

where $U \in \mathbb{R}^{n \times n}$ and $V \in \mathbb{R}^{m \times m}$ are orthogonal matrices, and $S \in \mathbb{R}^{n \times m}$ is "diagonal" with the nonnegative entries $\sigma_1 \geq \sigma_2 \geq \cdots \geq 0$ arranged along the main diagonal and where all other entries are equal to zero [16, 6]. Let $\tilde{U}$ represent the $n \times k$ submatrix formed from the first $k$ columns of $U$, and let $\tilde{V}$ represent the $m \times k$ submatrix formed from the first $k$ columns of $V$, while $\tilde{S}$ is the $k \times k$ submatrix formed from the first $k$ rows and columns of $S$. Then, assuming $\sigma_k > 0$, a rank-$k$ approximation to $A$ is given by

$$\tilde{U}\tilde{S}\tilde{V}^T. \tag{3}$$

It follows that $W = \tilde{U}\tilde{S}$ and $H = V^T$. Finally, note that if we were to subtract the mean from each row of $A$ before computing the SVD, then columns of $W = \tilde{U}\tilde{S}$ would be scalar multiples of the first $k$ principal components determined via principal component analysis (PCA) applied to the set of $m$ points in $n$ [5].

*2.3.2. Low-rank approximations via NMF*

The squared-error of the approximation produced by NMF will generally be larger than the squared-error of the approximation produced by the SVD. This is a consequence of the Eckart-Young theorem, which asserts that the approximation to $A$ obtained by truncating the SVD of $A$ is the rank-$k$ approximation with minimal squared-error [16, 6]. We note that the problem of finding the NMF is a non-convex problem that can be interpreted geometrically as finding a simplicial cone in the positive orthant for the set of column vectors of a matrix [7, 4]. Nevertheless, as mentioned above, we consider the NMF for the following three reasons: 1) it is capable of providing a low-rank approximation of an image, 2) it preserves the sign-structure of the original dataset, and 3) it produces a *parts-based decomposition* of the input image. We investigate the parts that are recovered by NMF in Section 4.1.1

To compute the factorization with NMF, the matrix $A \in \mathbb{R}^{n \times m}$ is decomposed as $A \approx WH$, where $W \in \mathbb{R}^{n \times k}$ and $H \in \mathbb{R}^{k \times m}$ have only nonnegative entries. In practice, the matrices $W$ and $H$ are obtained as iterates of the optimization

$$\min_{(W \in \mathbb{R}_+^{n \times k}, H \in \mathbb{R}_+^{k \times m})} \|A - WH\|_F, \tag{4}$$

where $\mathbb{R}_+$ represents the nonnegative real numbers, and $\|\cdot\|_F$ is the Frobenius norm. We use the multiplicative update algorithm described in [8]. We initiate the multiplicative update algorithm with random matrices for $W$ and $H$. Other authors have shown that alternate initializations such as spherical k-means improve the performance of the algorithm [17]. The derivation of the multiplicative update involves the use of Lagrange multipliers with nonnegative constraints. After the optimization converges to estimates $\tilde{W}$ and $\tilde{H}$, we define the factorization obtained via NMF has $T_k(A) = \tilde{W}\tilde{H}$.

## 3. The proposed algorithm

The main algorithm studied in this paper consists of three steps. First, we encode the image by defining a transformed matrix, $\hat{A}$, by partitioning $A$ into blocks. These blocks are then reorganizing into columns of $\hat{A}$ using lexicographic ordering of each block. Second, we compute a decomposition for $\hat{A}$ of the form (1). Last, our decoding scheme involves inverting the operation of creating columns from patches in order to obtain an approximation to $A$. We detail these steps below and in figure 1.

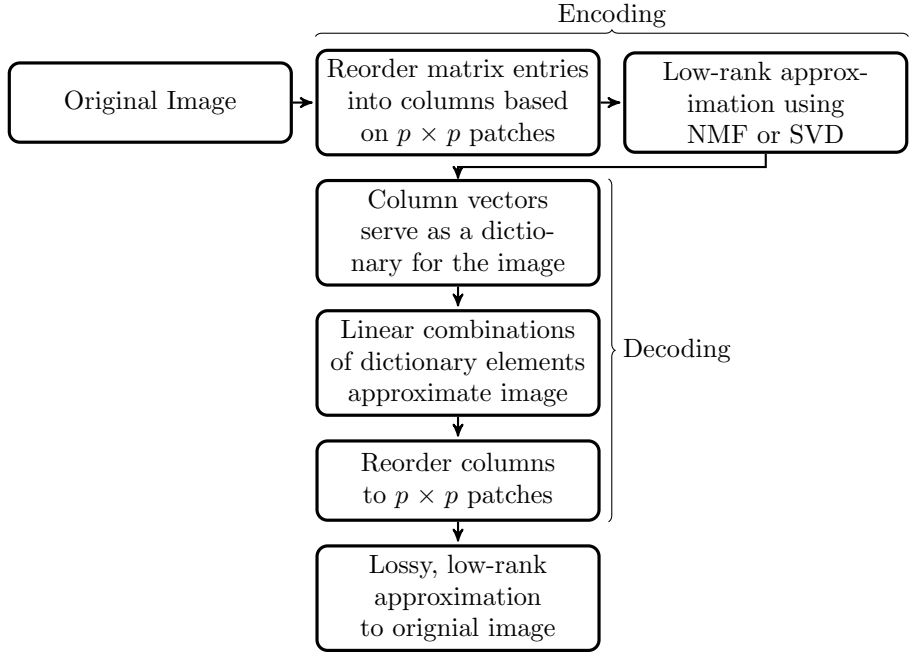

Figure 1: Encoding an image by reordering pixel entries based on patches.

3.1. Reordering pixels

Before compressing a matrix using a factorization of the form (1), our algorithm first reorders the entries of $A$. Specifically, $p$ is chosen to be an integer factor of $n$ and $m$. Then, $A$ is partitioned into non-overlapping, contiguous patches of size $p \times p$ pixels. This results in $nm/p^2$ different patches. Each patch is then shaped into a column vector in $\mathbb{R}^{p^2}$ by ordering the $p^2$ values *lexicographically* [2] as components of the column vector. Finally, we organize the resulting column vectors into a $p^2 \times \frac{nm}{p^2}$ matrix, $\hat{A}$. This procedure defines the operator

$$S_p : \mathbb{R}^{n \times m} \to \mathbb{R}^{p^2 \times \frac{nm}{p^2}},$$

---

[2] Lexicographic ordering traverses the pixels of each patch row from left to right and each subsequent patch row from top to bottom.

where $S_p(A) \equiv \hat{A}$. Note that the procedure that created $\hat{A}$ can be reversed to recover the original matrix $A$, thereby defining the inverse operation

$$S_p^{-1} : \mathbb{R}^{p^2 \times \frac{nm}{p^2}} \to \mathbb{R}^{n \times m}.$$

An example of $A$ and $S_p(A)$ is given in Figure 2, using $m = n = 256$ and $p = 16$. Notice that when $A$ is square and when $p = \sqrt{n}$, as in the example of Figure 2, the matrices $A$ and $\hat{A}$ are the same size. Also notice that the subtle breaks in the continuity of intensity that extend horizontally across $\hat{A}$ are caused by the left and right edges of each block after being shaped into column vectors. Other authors in [10, 9, 13] include ordering schemes of the vectors associated with patches. In our approach, the ordering of the patches is based solely on the left to right, then top to bottom traversal of the pixels of the image.

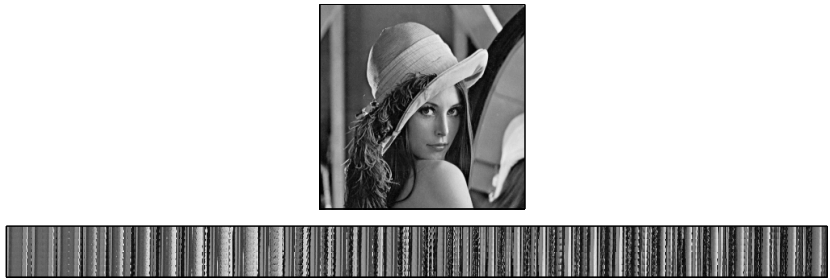

Figure 2: Top: An example of the $A$ matrix, representing a $256 \times 256$ image. Bottom: The associated reordered matrix $S_8(A)$. Note that $S_8(A)$ is $64 \times 1024$ dimensional.

*3.2. Encoding, decoding, and dictionaries*

Given $\hat{A}$, we define $\hat{T}_{\hat{k}} : \mathbb{R}^{p^2 \times \frac{nm}{p^2}} \to \mathbb{R}^{p^2 \times \frac{nm}{p^2}}$ as the operator which returns the rank-$\hat{k}$ approximation to $\hat{A}$, similar to the decomposition given in (1). That is

$$\hat{T}_{\hat{k}}(\hat{A}) = \hat{W} \hat{H}, \tag{5}$$

where $\hat{W} \in \mathbb{R}^{p^2 \times \hat{k}}$ and $H \in \mathbb{R}^{\hat{k} \times \frac{nm}{p^2}}$. By encoding the values of $\hat{W}$ and $\hat{H}$, the memory footprint of the decomposition in (5) is $\hat{k}\left(p^2 + \frac{nm}{p^2}\right)$.

Finally, we define the decoded image as $Q_{\hat{k},p}(A)$, where the operator $Q_{\hat{k},p} : \mathbb{R}^{n \times m} \to \mathbb{R}^{n \times m}$ is defined as

$$Q_{\hat{k},p} = S_p^{-1} \circ \hat{T}_{\hat{k}} \circ S_p. \qquad (6)$$

The main matrix spaces and relevant operators are illustrated in the schematic of Figure 3.

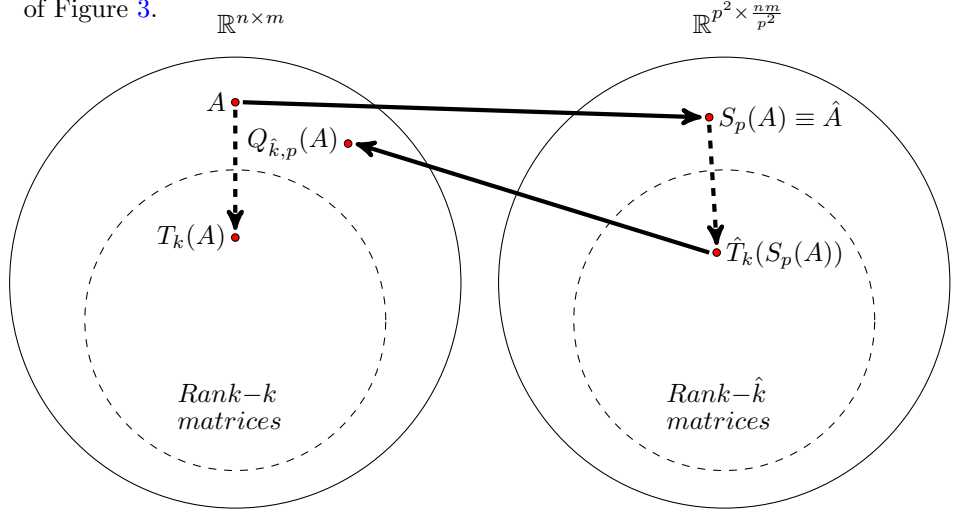

Figure 3: The main matrix spaces, and the relevant operators. Note that the reconstructed matrix $Q_{\hat{k},p}(A)$ is not necessarily rank $k$.

*3.2.1. Dictionaries from decompositions*

It is clear that the decomposition (5) approximates each column of $\hat{A}$ with a linear combination of columns of $\hat{W}$. Since each column of $\hat{A}$ represents a patch extracted from the original image $A$, the columns of $\hat{W}$ can be thought of as dictionary elements used to define all patches from the original image. When NMF is used as the low-rank approximation producing (5), this interpretation means that the dictionary elements are like parts of the image. See Section 4.1.1 for examples of the elements of $\hat{W}$ reshaped as patches.

## 4. Analysis of the proposed algorithm

In this section we explore the characteristics of the proposed algorithm, as well as evaluate its performance. In particular, we provide a qualitative assessment of the different factorizations and reconstructed images. In addition, we also quantitatively analyze the memory footprint of each compression strategy, and investigate the impact it has on the quality of the reconstruction.

*4.1. Qualitative assessment*

Examples of $T_k(A)$ are given in Figure 4. The SVD and the NMF are used to compute the factorization in (5), which yields the reconstructions in the top and bottom rows of images, respectively. Since the Eckart-Young theorem asserts that the approximation to $A$ obtained with the SVD is the rank-$k$ approximation with the smallest error in the Frobenius-norm, it is expected that the SVD examples exhibit higher visual quality than those that use NMF. In Figure 5 we also include magnified views of portions of the reconstructions using the SVD and NMF on the columns of the images in figure 4.

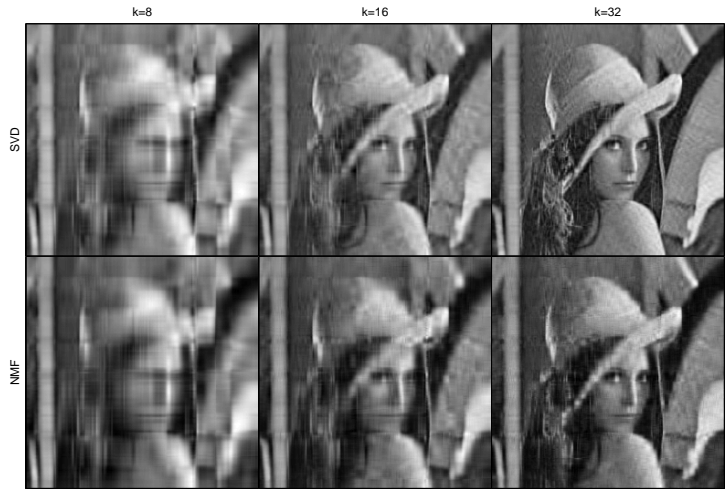

Figure 4: Various rank-$k$ reconstructions $T_k(A)$ of the original $256 \times 256$ image matrix $A$, using the SVD (top row) and NMF (bottom row). The parameter $k$ changes from left-to-right: $k = 8$, $k = 16$, and $k = 32$.

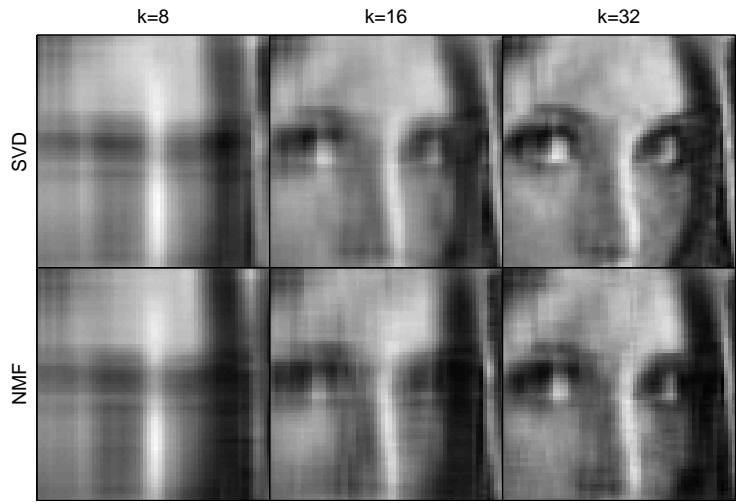

Figure 5: Magnified views of various rank-$k$ reconstructions $T_k(A)$ of Figure 4, using the SVD (top row) and NMF (bottom row). The parameter $k$ changes from left-to-right: $k = 8$, $k = 16$, and $k = 32$.

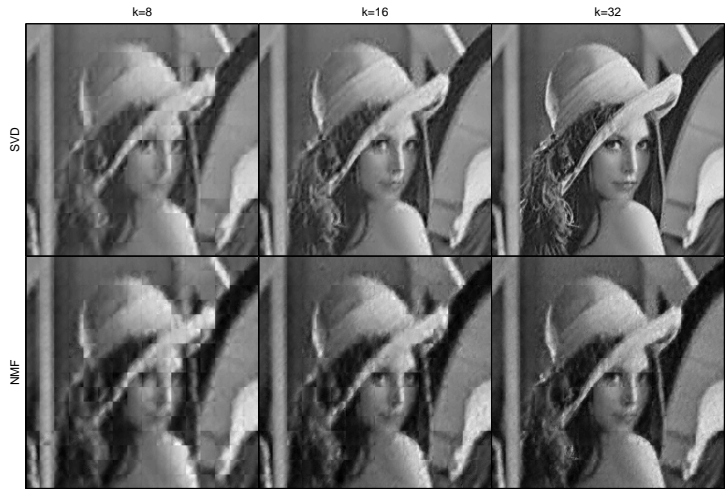

Figure 6: Several examples of $Q_{\hat{k},p}$, using $\hat{k} = 8, 16, 32$ from left-to-right. The SVD is used for the reconstructions in the top row, while NMF is used for the reconstructions in the bottom row.

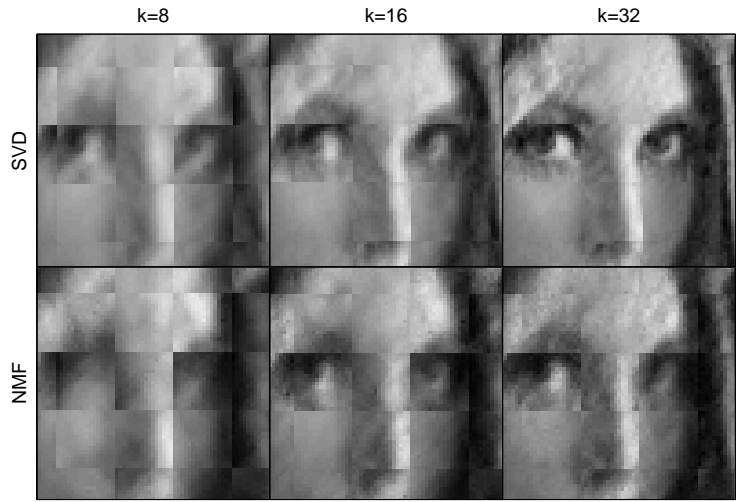

Figure 7: Magnified examples of Figure 6 using the proposed approach $Q_{\hat{k},p}$, using $\hat{k} = 8, 16, 32$ from left-to-right. The SVD is used for the reconstructions in the top row, while NMF is used for the reconstructions in the bottom row.

Similarly, examples of $Q_{\hat{k},p}$ are given in Figure 6 using the same image of size $256 \times 256$ and a fixed value of $p = 16$. We notice that the reconstructed examples of $Q_{\hat{k},p}$ in Figure 6 are visually better in general than their counterparts in Figure 4. We also observe that artifacts of the patch-grid are evident in the reconstructions of $Q_{\hat{k},p}$, especially for smaller $k$. This is also illustrated in the magnified views shown in Figure 7.

### 4.1.1. The columns of $\hat{W}$s dictionary elements

Although the NMF approximations are not expected to be better than the SVD as pointed out in Section 3.2, we do expect that NMF will produce a matrix $\hat{W}$ in the product (5) having a parts-based structure. This follows from the fact that each column of $\hat{A}$ is constructed using a patch from $A$ that has been re-shaped into a column vector. In addition, the approximation (5) represents each column of $\hat{A}$ as a linear combination of columns of $\hat{W}$. Therefore, it follows that the columns of $\hat{W}$ are effectively elements of a dictionary that are used to approximate each of the patches from the original image $A$. This is similar to the application of NMF to a library of images in [8]. However, in this paper, rather than finding parts that belong to many images in a set, we find parts that belong to a single image in the form of localized patches.

We illustrate this observation below, where we visualize elements of the SVD and NMF dictionaries that are computed in (5). Examples of the patches from the original image, which are ultimately organized as columns of $\hat{A}$, are shown in Figure 8.

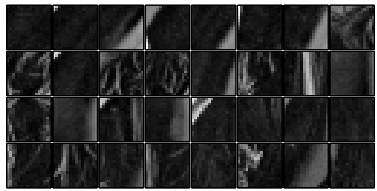

Figure 8: Examples of several patches extracted from the original image (left) of Figure 2.

We provide examples of the dictionary elements from the associated matrix $\hat{W}$ of rank $k = 32$. Specifically, we reshape the first 32 columns of $\hat{W}$ into $p \times p$ patches of the $4 \times 8$ array of patches in Figure 9. In particular, the first eight columns of $\hat{W}$ are represented as patches in the first row, the second eight columns are represented as patches in the second row, and so on. Two such arrays are shown in Figure 9. The left $4 \times 8$ array of patches are from the SVD dictionary, while the right $4 \times 8$ array of patches are from the NMF dictionary.

One of the most noticeable aspects of the elements in the SVD dictionary is the fact that the sign structure of its entries is not preserved, as indicated by the light blue pixels. Since the black pixels represent the zero value, each element has global support over the entire $p \times p$ patch. In addition, we observe that the first, third, and fourth elements represent slow changes from positive to negative in from the left-to-right, top-to-bottom, and diagonal directions, respectively.

In contrast, the elements in the NMF dictionary have local support and are nonzero only over a relative few of the pixels in the $p \times p$ patches. Also, due to the parts-based-decomposition produced by NMF, we see that the elements of the NMF dictionary resemble edges and curves more closely than the elements in the SVD dictionary.

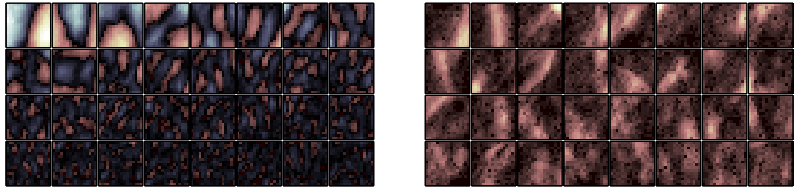

Figure 9: Examples of the dictionary elements used to create the reconstructions in Figure 6. The left panel shows a color image of the set of 32 vectors associated with the SVD dictionary while the right panel shows a color image for the set of 32 vectors associated with the NMF dictionary. Pixel values of zero are represented with black, positive values with pink, and negative with light blue.

## 4.2. Comparing memory footprints

In this section, we compare the amount of compression achieved by the proposed algorithm using the memory footprint described in Section 2.2. Table 1 shows the memory footprint of the original image matrix $A$, as well as the reconstructions $T_k(A)$ and $Q_{\hat{k},p}(A)$.

Table 1: Memory footprints of relevant matrices

| Matrix | Memory footprint |
|---|---|
| $A$ | $nm$ |
| $T_k(A)$ | $k(n+m)$ |
| $Q_{\hat{k},p}(A)$ | $\hat{k}\left(p^2 + \frac{nm}{p^2}\right)$ |

There are several things to observe based on Table 1, and the examples that have been presented so far.

1. Notice that even without reordering the matrix, the amount of compression achieved by $T_k(A)$ is still significant. For example, the reconstruction $T_k(A)$ in the top-right of Figure 4, where $n = m = 256$ and $k = 32$, uses roughly 75% less information than the original image in the left of Figure 2. In addition, there are minor artifacts associated with the reconstructions using this approach.

2. If $\hat{k} = k$, and if $p = \sqrt{n}$ or if $p = \sqrt{m}$, then the memory footprint of $Q_{\hat{k},p}(A)$ is equal to the memory footprint of $T_k(A)$. For example, the memory footprint of $Q_{\hat{k},p}(A)$ in the top-right of Figure 6 is the same as for $T_k(A)$, yet the visual quality of $Q_{\hat{k},p}(A)$ is better.

3. More generally, note that the memory footprint of $Q_{\hat{k},p^*}(A)$ is the product of $\hat{k}$ and the function $g(p) = p^2 + \frac{nm}{p^2}$. Since $n$ and $m$ are fixed by the size of the input image, the function $g(p)$ is minimized when $p = (nm)^{1/4}$, and has the minimum value $2\sqrt{nm}$. In other words, the memory footprint of $Q_{\hat{k},p}(A)$ is smallest when $p = (nm)^{1/4}$, and is equal to $2\hat{k}\sqrt{nm}$.

4. We denote the memory footprint of $Q_{\hat{k},p}(A)$ when $p$ is the optimal value $p^* \triangleq (nm)^{1/4}$ as $\mu(Q_{\hat{k},p^*}(A))$. Similarly, we denote the memory footprint of $T_k(A)$ as $\mu(T_k(A))$. Since $2\sqrt{nm} \leq n + m$,[3] we get the inequality

$$\mu(Q_{\hat{k},p^*}(A)) \leq \frac{\hat{k}}{k}\mu\left(T_k(A)\right), \tag{7}$$

with equality holding only if $n = m$. We note that while the parameter $k$ used to define the operator $T_k$ generally satisfies $1 \leq k \leq \min(n,m)$, the parameter $\hat{k}$ satisfies $1 \leq \hat{k} \leq \min\left(p^2, \frac{nm}{p^2}\right)$. Therefore, for smaller values of $p$, $\hat{k}$ is likely less than $k$, which makes the ratio in (7) less than one.

A graph of $\mu(Q_{\hat{k},p}(A))$ for $n = m = 256$ and $\hat{k} = k = 32$ is given in the left panel of Figure 10 as a function of patch size $p$.[4] Notice that $\mu(Q_{\hat{k},p}(A))$ is smallest when $p = p^* = 16$. This indicates that the patch size of $16 \times 16$ is optimum in minimizing the memory footprint. The right panel of Figure 10 shows the same curve, but using parameters $n = 900$ and $m = 1600$.

---

[3]This result is based on the fact the the geometric mean is at most the arithmetic mean.
[4]The value of $k$ is essentially a multiplicative scaling, and doesn't affect the shape of the curve.

Although the image is not square, we continue to define square patches of size $p \times p$ on the image. Also, since $n \neq m$, the quantity $\mu(Q_{\hat{k},p^*}(A))$ is strictly less than $\mu(T_k(A))$, according to (7). This can be observed in the right-hand-side of Figure 10, where the green-curve passes under the constant horizontal curve around $p = (nm)^{1/4} \approx 34.64$.

In practice, we will not choose $\hat{k} = k$. In fact, it will be impossible if $\min(p^2, \frac{nm}{p^2}) < k$, since $\hat{k}$ is bounded above by the rank of $\hat{A}$. This means that the memory footprint of $Q_{\hat{k},p}(A)$ will often be much smaller than the memory footprint of $T_k(A)$ when $\hat{k} < k$ for most values of $p$. We explore the relationship between $p$, $k$, and $\hat{k}$ in Section 4.3 and Figures 12, 13 and 14. There, we demonstrate that $\hat{k}$ can be chosen to satisfy $\hat{k} < k$ in order to keep the memory footprint of $Q_{\hat{k},p}(A)$ comparable to the memory footprint of $T_k(A)$.

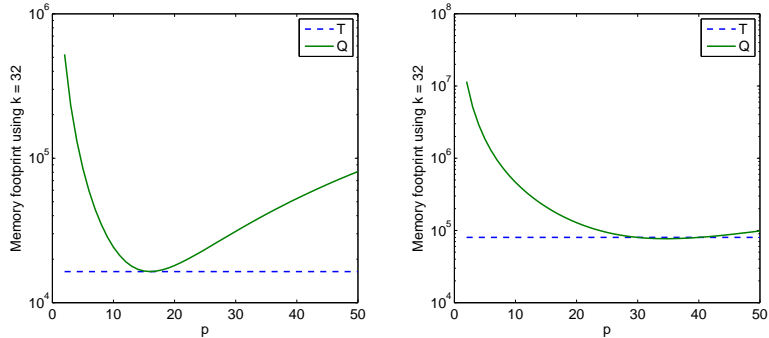

Figure 10: Memory footprints of $T_k$ (dashed curve), and $Q_{k,p}(A)$ as a function of $p$ (solid curve). The values $n = m = 256$ for the curves on the left, and $n = 900$ and $m = 1600$ for the curves on the right.

### 4.3. Quality as a function of memory footprint

In this section, we report the results of experiments where we compare the quality of the reconstructions produced via the proposed algorithm. Specifically, we compare the mean-squared-error of $T_k(A)$ and $Q_{\hat{k},p}(A)$ when the memory footprint of $T_k(A)$ is bounded from below by the memory footprint of $Q_{\hat{k},p}(A)$.

The heart of our results lie in Figures 12, 13 and 14 where we plot the PSNR as a function of the memory footprint. The PSNR is defined as

$$\text{PSNR} = 10 \log_{10}\left(\frac{1}{e^2}\right),$$

where $e^2$ is the mean-squared-error between the original image $A$ and the approximations. The two figures on the left in Figure 12 are computed using the SVD, while the two figures on the right in Figure 12 are computed using NMF.

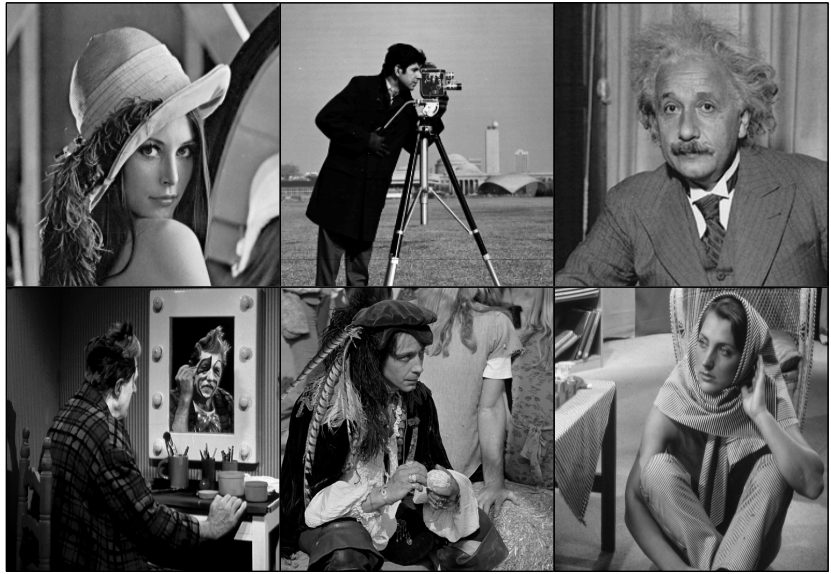

Figure 11: The six test images used for the evaluation of the algorithm. From left to right, they are: Lena, Cameraman, Einstein, Clown, Man and Barbara.

To create these curves, we first fix $A$ to be one of the test images from Figure 11, which also fixes $n = m = 256$. Then, we choose $k$ from a set of values between 2 and 50. With $k$ chosen, we then choose $p$ from the set $\{4, 8, 16, 32, 64\}$. This enables us to finally choose the largest integer $\hat{k}$ such that

$$\hat{k}\left(p^2 + \frac{nm}{p^2}\right) \leq k(n+m).$$

With the parameters chosen to ensure that the memory footprints are comparable, we compute the mean-squared-errors of each $T_k(A)$ and $Q_{\hat{k},p}(A)$. For example, we compute the mean-squared-error between $A$ and $T_k(A)$ to be

$$\frac{1}{nm}\|A - T_k(A)\|_F,$$

where $\|\cdot\|_F$ is the Frobenius norm. Similarly, we compute the mean-squared error between $A$ and $Q_{\hat{k},p}(A)$ to be

$$\frac{1}{nm}\|A - Q_{\hat{k},p}(A)\|_F.$$

Figure 3 provides an illustration of these error estimates with regard to their spaces and low-rank approximations of the original matrix $A$.

Figure 12 demonstrates that for the smallest memory footprints (largest compression rates), the best reconstructions of the five test images are achieved using the proposed algorithm with $p = 8$. We also see variations in the smoothness of the curves associated with the application of NMF. Recall that the NMF problem has many solutions and that because of the nature of the multiplicative update algorithm, local minima are encountered. Whereas the SVD finds the absolute minimum for each patch size resulting in a smooth curve. The benefit associated with using the NMF algorithm lies in the fact that the dictionary of vectors which is generated, has a parts-based, localized representation.

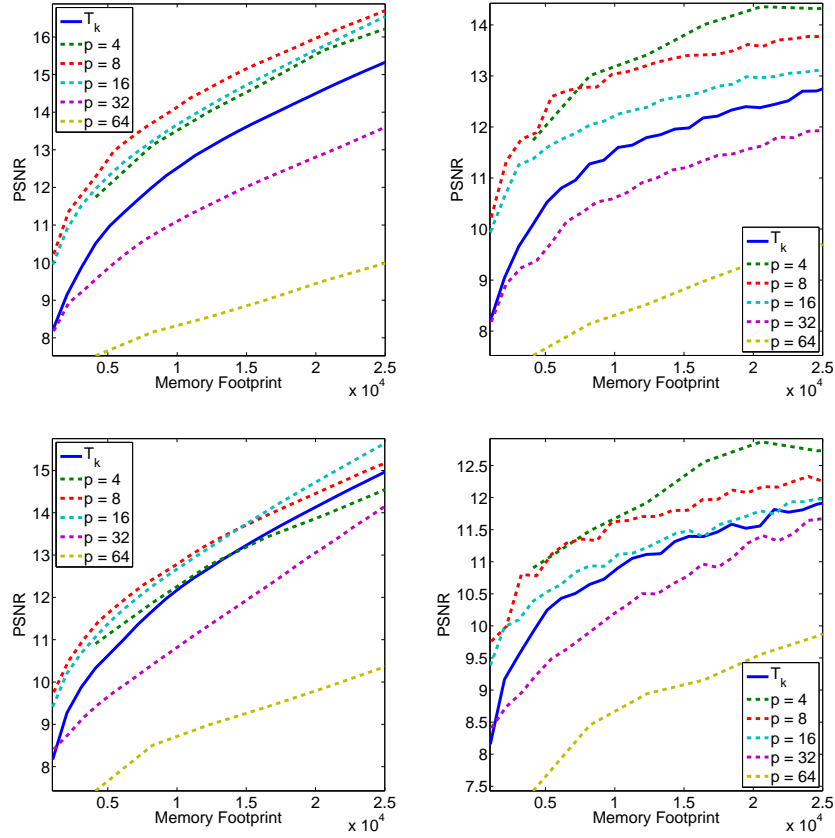

Figure 12: The PSNR as a function of memory footprint for $T_k(A)$ (solid blue curves), and $Q_{\hat{k},p}(A)$ (dashed curves) for various values of $p$. The two rows are generated using Lena and Cameraman, respectively. The left column is generated using the SVD to compute the factorization in (1). The two figures on the right are generated using NMF. Ideally, the highest PSNR with the lowest footprint is most desirable. Note also that variations in the curves associated with NMF is due to solving the non-convex optimization problem at every step of the multiplicative update algorithm.

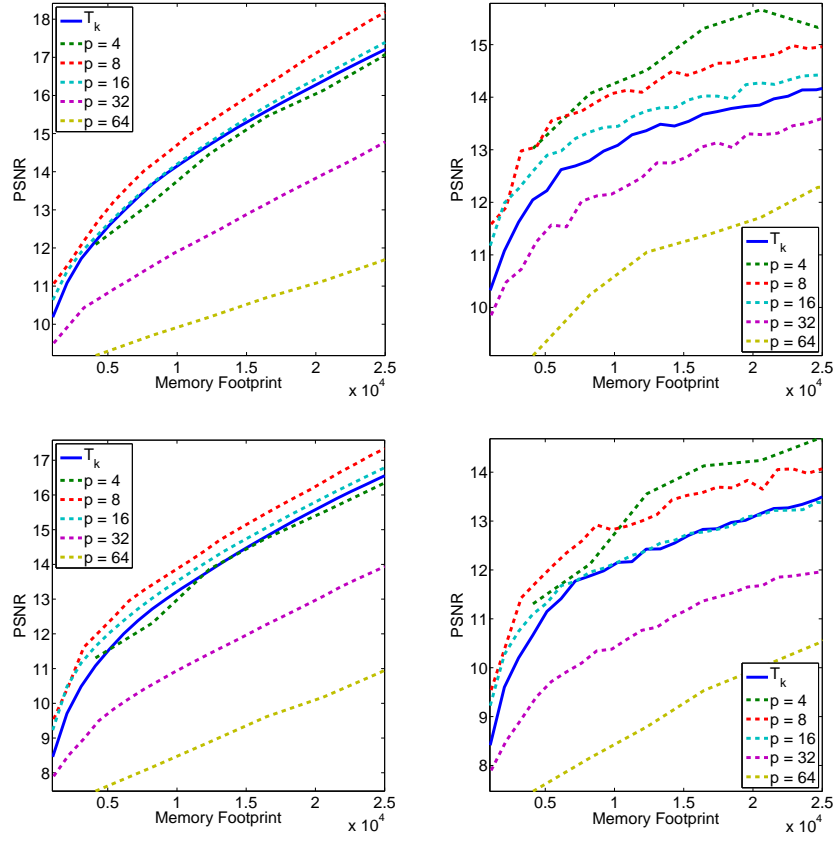

Figure 13: The PSNR as a function of memory footprint for $T_k(A)$ (solid blue curves), and $Q_{\hat{k},p}(A)$ (dashed curves) for various values of $p$. The two rows are generated using Einstein and Clown, respectively. The left colunm of the figure is generated using the SVD to compute the factorization in (1). The two figures on the right are generated using NMF.

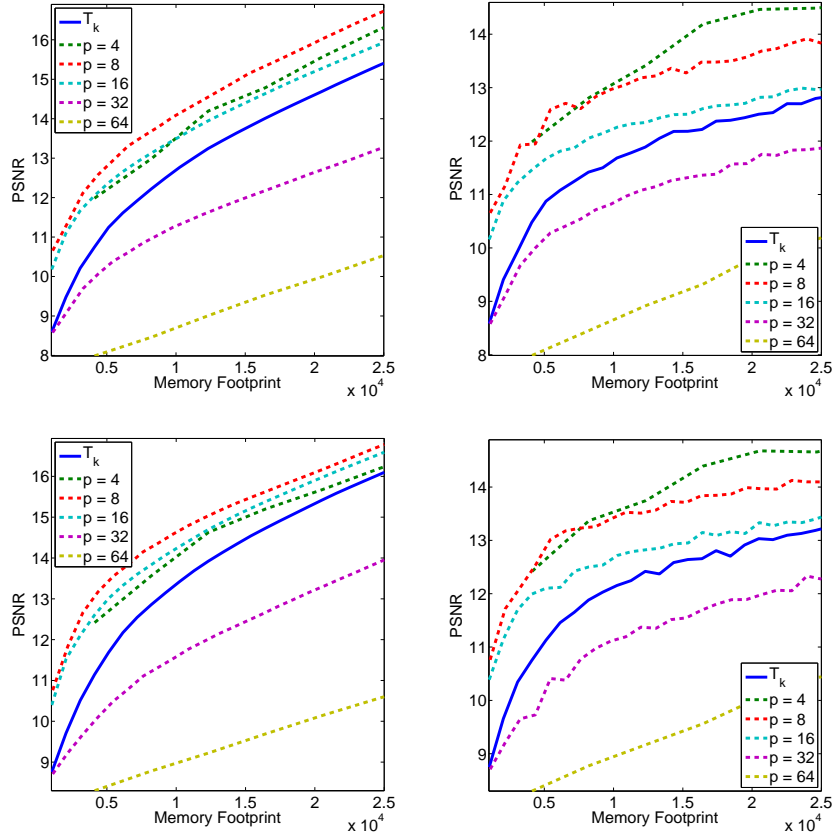

Figure 14: The PSNR as a function of memory footprint for $T_k(A)$ (solid blue curves), and $Q_{\hat{k},p}(A)$ (dashed curves) for various values of $p$. The two rows are generated using Man and Barbara, respectively. The left colunm of the figure is generated using the SVD to compute the factorization in (1). The two figures on the right are generated using NMF.

We also use the Structural Similiarity Index (SSIM) [15] as an objective measure of how well the approximated image agrees with the original image. The SSIM is a well known alternative to providing an assessment of the preserved structures within an image as opposed to the MSE, which determines agreement of two images at the level of individual pixels. The SSIM separates the task of similarity measurement into three components: luminance, structure and contrast. These components are well defined and are combined in order to provide information about the structure of an image based on patches. The SSIM equation is as follows:

$$SSIM(x,y) = \frac{(2\mu_x\mu_y + C_1)(2\sigma_{xy} + C_2)}{(\mu_x^2 + \mu_y^2 + C_1)(\sigma_x^2 + \sigma_y^2 + C_2)} \qquad (8)$$

In equation (8), $\mu_x$ represents the mean intensity for a path from image 'x', and $\sigma_x$ represents the normalized unbiasd estimate of the intensities. In short, the estimate of the SSIM of two images satisfy the following conditions:

1. Symmetry: $S(x,y) = S(y,x)$;

2. Boundedness: $S(x,y) \leq 1$;

3. Unique maximum: $S(x,y) = 1$ if and only if $\mathbf{x} = \mathbf{y}$ ($x_i = y_i$, for all $i = 1, 2, ..., N$);

In table 2 we show the mean SSIM values for the test images when the standard SVD and NMF algorithms are applied using $T_k(A)$ and $Q_{\hat{k},p}$. The mean SSIM (MSSIM) is calculated for the purpose of achieving an overall quality measure for an image. We define the MSSIM as follows:

$$MSSIM(\mathbf{X}, \mathbf{Y}) = \frac{1}{M} \sum_{j=1}^{M} SSIM(\mathbf{x}_j, \mathbf{y}_j), \qquad (9)$$

where $\mathbf{X}$ and $\mathbf{Y}$ are the original and approximated images, respectively; $\mathbf{x}_j$ and $\mathbf{y}_j$ are the image contents at the j-th window; and $M$ is the number of windows in the image. We define the local window to be a window where local statistics are measured. In this application, we use a two-dimensional Gaussian lowpass filter with a standard deviation of 1.5 pixels.

Table 2: Mean SSIM of relevant images is shown. Using a patch size of $p = 16$, images designated with 'p' indicate our proposed patch-based algorithm. Two identical images would have a structural similarity index of 1. Take special note of the mean SSIM scores for the Einstein and Clown standard and patch-based algorithms for k=8.

|             | **SVD** | | | **NMF** | | |
| :---: | :---: | :---: | :---: | :---: | :---: | :---: |
| **Image** | **k=8** | **k=16** | **k=32** | **k=8** | **k=16** | **k=32** |
| Lena        | 0.5873 | 0.6787 | 0.7892 | 0.5709 | 0.6316 | 0.6726 |
| Lena-p      | 0.6682 | 0.7696 | 0.8583 | 0.6169 | 0.6650 | 0.7094 |
| Cameraman   | 0.5576 | 0.6579 | 0.7671 | 0.5608 | 0.5830 | 0.6257 |
| Cameraman-p | 0.6421 | 0.7259 | 0.8138 | 0.5739 | 0.5982 | 0.6273 |
| Einstein    | 0.6859 | 0.7618 | 0.8647 | 0.6563 | 0.6850 | 0.7246 |
| Einstein-p  | 0.6917 | 0.7971 | 0.8889 | 0.6450 | 0.6893 | 0.7291 |
| Clown       | 0.6088 | 0.7241 | 0.8425 | 0.6127 | 0.6815 | 0.7472 |
| Clown-p     | 0.6392 | 0.7664 | 0.8704 | 0.6057 | 0.6836 | 0.7476 |
| Man         | 0.5159 | 0.6375 | 0.7737 | 0.4964 | 0.5664 | 0.6384 |
| Man-p       | 0.5911 | 0.7060 | 0.8205 | 0.5514 | 0.5956 | 0.6576 |
| Barbara     | 0.5894 | 0.7059 | 0.8167 | 0.5713 | 0.6338 | 0.6820 |
| Barbara-p   | 0.6665 | 0.7656 | 0.8556 | 0.6097 | 0.6575 | 0.7013 |

Finally, in Figure 15 we illustrate the dictionaries which were generated using a rank of $k = 32$ with $p = 16$. Note the noticeable structure in the NMF patches using our proposed algorithm.

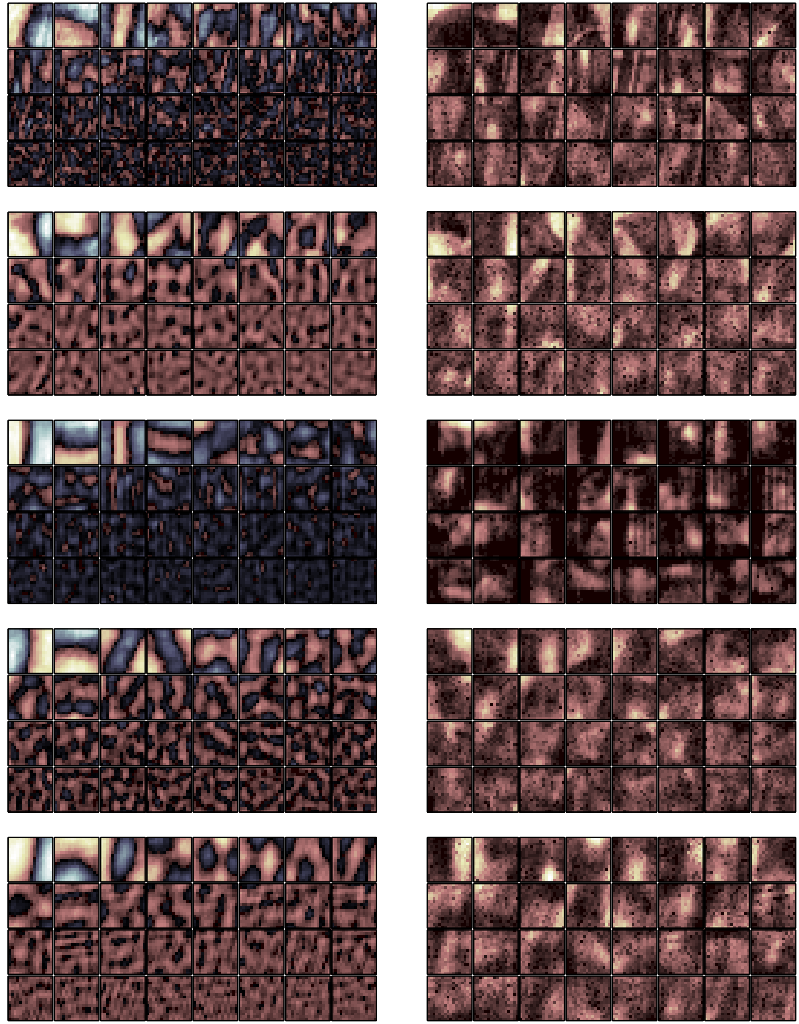

Figure 15: Examples of the dictionary elements used to create the reconstructions in Figure 6. The dictionaries generated from the top row to the bottom, are as follows: Cameraman, Einstein, Clown, Man and Barbara. The left panel shows a color image of the set of 32 vectors associated with the SVD dictionary while the right panel shows a color image for the set of 32 vectors associated with the NMF dictionary. Pixel values of zero are represented with black, positive values with pink, and negative with light blue.

## 5. Discussion

Given a mechanism for producing a low-rank approximation to an image matrix, our experiments demonstrate that a higher-quality reconstruction can be obtained by applying the same mechanism to a reordered version of the image pixels that is based on patches. Our approach also yielded a methodology for creating nonnegative, structured dictionaries for single images, which are localized and parts-based. In addition to producing higher-quality reconstructions at a given compression rate, the proposed algorithm is generally faster when $p$ is small. Below, we detail the computational cost of the proposed algorithm, provide some intuition behind choosing the size of $p$, and suggest several open questions based on this paper.

### 5.1. Computation Considerations

Our approach relies on the ability to reorder the image and invert this operation, as well. Note that the reordering operation (and its inverse) requires $O(nm)$ operations. Also, when using the SVD, computing $T_k(A)$ and $\hat{T}_{\hat{k}}(\hat{A})$ requires $O(mn^3)$ and $O(mnp^4)$, respectively. This follows from the observation that one can compute the reduced SVD of an $n \times m$ matrix to within machine tolerance in approximately $4mn^3 - 4n^4/3$ operations (assuming $n < m$) [16, 6]. So, computing $\hat{T}_{\hat{k}}(\hat{A})$ requires approximately $\frac{p^4}{n^2}$ as much time computing $T_k(A)$, which is substantial savings in the ideal situation that $p$ is a small fraction of $n$. Indeed, experiments show that reordering the image (and inverting this operation) is negligible to computing $T_k(A)$ and $\hat{T}_{\hat{k}}(\hat{A})$. This was also observed when using NMF instead of the SVD.

## 5.2. MSSIM Considerations

The mean SSIM (MSSIM) is an insightful quantity and showed that in 34 of the 36 cases considered, the performance of the patched-based NMF and SVD algorithms was better than the non-patched based approaches. The reason for these exceptions is not yet clear. However, we note that the images where these exceptions occured, have vertical structures in their backgrounds. It is conceivable that the non-patched algorithms create vertical structures as a crude approximation for the low rank case of k=8. The results indicate that for both the SVD and NMF implementations, our approach yielded a higher mean SSIM in the test images for all ranks. This provides additional validation of our approach.

## 5.3. Parameter Selection

Our results establish that the quality of the reconstruction, the amount of information required, and the computation savings of the proposed algorithm is dependent on the size of the image patches. Our experiments suggest that medium-sized patches (e.g. $8 \times 8$) often lead to higher-quality reconstructions. Several of the ideal-sized patches extracted from our test images are exhibited in Figure 8. We observe that the patches contain relatively little information, compared to the entire image. However, the example patches do show complex image features, such as edges, corners, shading, and textures. This suggest that $p$ should be chosen based on the natural scale of the local parts of the image. That is, $p$ should be large enough to capture salient features such as edges and corners, but not too large that a single patch captures more macroscopic content in the image, as illustrated in Figure 12.

## 5.4. Nature of Dictionaries

In Figure 15 we compare the nature of the dictionary elements which are generated using standard SVD and NMF without the use of patches and their patch implementations. What the data revealed is the fact that the dictionary elements do seem to have nonnegative elements that reflect the structures within the image. Although structure within the dictionaries is present, the diictionary elements were not very sparse. Exploring techniques to increase the sparsity of the approach seems to be a viable path to pursue. In addition, investigating the impact of the algorithm on single image, repeated patterns may offer some additonal insight.

*5.5. Assumptions and Open Questions*

Our approach is based on the assumption that the matrix being compressed represents a natural image that contains coherent information about the local, salient features of the image. Indeed, preliminary experiments suggest that when the matrix to be compressed is not a natural image (e.g. an image of random pixel values), the quality of the reconstruction $Q_{\hat{k},p}$ is at most equal to the quality of the reconstruction $T_k(A)$.

Our approach also assumes that the parameter $p$ can be chosen as an integer factor of both $m$ and $n$. As mentioned above, ideally $p$ is chosen based on the natural scale of features in the image, rather than on the size of the image. Also, since our patches are only one size, it would be desirable to incorporate patches of different sizes to account for the multi-scale nature of features in natural images.

We also propose pursuing the use of the algorithm on repeated patterns. An open question relates to the effected of patches that are repeated with the same orientation. This is not dissimilar to the original application of NMF on cannonical images.

**Acknowledgments**

The authors acknowledge the insightful comments, suggestions and recommendations of the referees on a previous version of this manuscript.

James H. Curry wishes to gratefully acknowledge that a portion of this work was carried out during his appointment as Program Director in the Division of Mathematical Sciences at the National Science Foundation.

**Vitae**

Richard M. Charles received his MS in Applied Mathematics from the University of Colorado at Boulder in 1995. His research interests include the modeling and analysis of large datasets using Nonnegative Matrix Factorization and Radial Basis Functions.

Kye Taylor received his Ph.D. from the University of Colorado at Boulder in 2011. He currently is a lecturer of mathematics at Tufts University.

Dr. James H. Curry, is Professor of Applied Mathematics at the University of Colorado-Boulder, since 1988. His research interest are dynamical systems, numerical linear algebra and image processing. Between 2003-2012 he served as Chair; and 2012-2014 he served as a Program Director in the Division of Mathematical Sciences at NSF.